# The Power of Transfer Learning in Agricultural Applications: AgriNet

**Zahraa Al Sahili[1*], Mariette Awad[1]**

[1]Department of Electrical and Computer Engineering, Maroun Semaan Faculty of Engineering, American University of Beirut, Beirut, Lebanon

**\* Correspondence:**

Zahraa Al Sahili

zma35@mail.aub.edu

**Abstract**

Advances in deep learning and transfer learning have paved the way for various automation classification tasks in agriculture, including plant diseases, pests, weeds, and plant species detection. However, agriculture automation still faces various challenges, such as the limited size of datasets and the absence of plant-domain-specific pretrained models. Domain specific pretrained models have shown state of art performance in various computer vision tasks including face recognition and medical imaging diagnosis. In this paper, we propose AgriNet dataset, a collection of 160k agricultural images from more than 19 geographical locations, several images captioning devices, and more than 423 classes of plant species and diseases. We also introduce AgriNet models, a set of pretrained models on five ImageNet architectures: VGG16, VGG19, Inception-v3, InceptionResNet-v2, and Xception. AgriNet-VGG19 achieved the highest classification accuracy of 94 % and the highest F1-score of 92%. Additionally, all proposed models were found to accurately classify the 423 classes of plant species, diseases, pests, and weeds with a minimum accuracy of 87% for the Inception-v3 model. Finally, experiments to evaluate of superiority of AgriNet models compared to ImageNet models were conducted on two external datasets: pest and plant diseases dataset from Bangladesh and a plant diseases dataset from Kashmir.

## 1    Introduction

The world population is expected to reach over 9 billion by 2050, which will require an increase in food production by 70% [1]. Considering scarcity of resources and climate change, intervention of artificial intelligence (AI) in agriculture is needed to overcome this challenge [2]. AI advantages can span from plant diseases detection, robotic weeds and pests control, to herbal discovery. Plant diseases are not only a risk for food security only, but they also have disastrous effects on smallholder farmers where pests and weeds can lead to the destruction of around 50% of the farm's plants [3]. Automated recognition of weeds, pests, and plant diseases can support smallholder farmers through free diagnosis services using mobile applications. Additionally, weed control robotics and sensor monitoring are another form of automation applied in regions with a limited number of agricultural expertise. Another important detection task is automated plant species recognition which is used in medical herbal research and in preventing extinction of non-discovered plant species [4].

Historically, the recognition task was relying on algorithms that needs handcrafted features, which were processed using relatively simple discriminative models such as linear classifiers or support vector machines (SVM) [5-8]. After being the leading algorithm in all computer vision tasks, deep learning (DL) has been widely used in agriculture research for plant classification tasks [5]. To achieve good accuracy, DL models need very large datasets for their requirements as data-hungry neural networks [5,8]. In the agricultural domain, datasets size is limited, so transfer learning would allow models reach higher accuracy without the need for more field data [9]. However, pretrained models used for transfer learning are not agriculture domain specific and were trained general computer vision datasets such as ImageNet. This creates a big challenge since convolutional models moves from low level features to higher level features and can lead to negative transfer [9]. For example, Yan et. al proposed transfer learning framework based on synthetic images to improve in-vitro soybean segmentation [12]. The proposed framework resulted in a precision improvement of 8% considering the data abundancy in soybean applications [12].

Another challenge is models' robustness which is affected by the type of agricultural data use. Mohanty et al. compared usage of AlexNet and Google LeNet pretrained models for 26 diseases and 14 crops species through PlantVillage dataset which constitutes of 54,306 lab images. The Google LeNet achieved the highest accuracy of 99.3% [10]. However, upon testing on trusted online sources, the accuracy dropped drastically to 31.4% [10]. After that Singh et al. introduced PlantDoc, a 2,598 field images dataset of 13 crops and 27 classes [11]. To classify the dataset's images, multiple experiments were done on both uncropped and cropped images. For the non-cropped images, using ImageNet architectures with PlantVillage weights [10] resulted in twice accuracy compared to using the same architectures but with ImageNet weights [11]. In the cropped dataset experiment, transfer learning on VGG16 architecture with ImageNet weights resulted in an accuracy of 44.52% compared to 60.42% accuracy when VGG16 was used with plant village weights [11].

Class imbalance degrades the performance of deep learning models on small classes including agricultural applications. For example, transfer learning was applied through ResNet50 architecture by Thapa et al. to detect two common apple diseases: apple scab and apple rust [13]. The accuracy obtained was 97% with an accuracy of only 51% for mixed diseases, which was caused by the small number of apples that have both diseases [13]. Additionally, Teimouri et al. used deep learning in the estimation of the weed growth stage [14]. The dataset of 9649 images for various weed species was classified from 1 to 9 growth stages [14]. Inception-v3 model was selected due to its good performance and low computational cost, and transfer learning was applied resulting in a 70% accuracy with a minimum accuracy of 46% for black-grass species that had the smallest set of images in the dataset [14].

Motivated to provide the agritech field with domain specific pretrained models that are robust and generalizable in various agricultural applications, the contributions of this work can be summarized as follows:

1- AgriNet dataset: a collection of 160k agricultural images from more than 19 geographical locations, several images captioning devices, and more than 423 classes of plant species and diseases.
2- AgriNet models: a set of pretrained models on five ImageNet architectures: VGG16, VGG19, Inception-v3, InceptionResNet-v2, and Xception and using AgriNet dataset. The proposed models are introduced to robustly classify the 423 classes of plant species, diseases, pests, and weeds with a minimum accuracy 94%, 92%, 89% ,90%, and 88% for each architecture respectively.

3- Pretraining using AgriNet models: transfer learning using AgriNet models compared to ImageNet models was evaluated using experiments on two agricultural datasets: pest and plant diseases dataset from Bangladesh and a plant diseases dataset from Kashmir.

## 2   Materials and Methods

### 2.1   Dataset

The AgriNet dataset is a collection of 160142 images belonging to 423 plant classes. The dataset was collected from 19 public datasets [15-33] geographically distributed between United States, Denmark, Australia, United Kingdom, Uganda, India, Brazil, Pakistan, and Taiwan. It includes field and lab images from different cameras and mobile devices, and it can perform multiple agricultural classification tasks, such as species, weed, pest, and plant diseases detection. Sample dataset images is displayed in Figure 1.

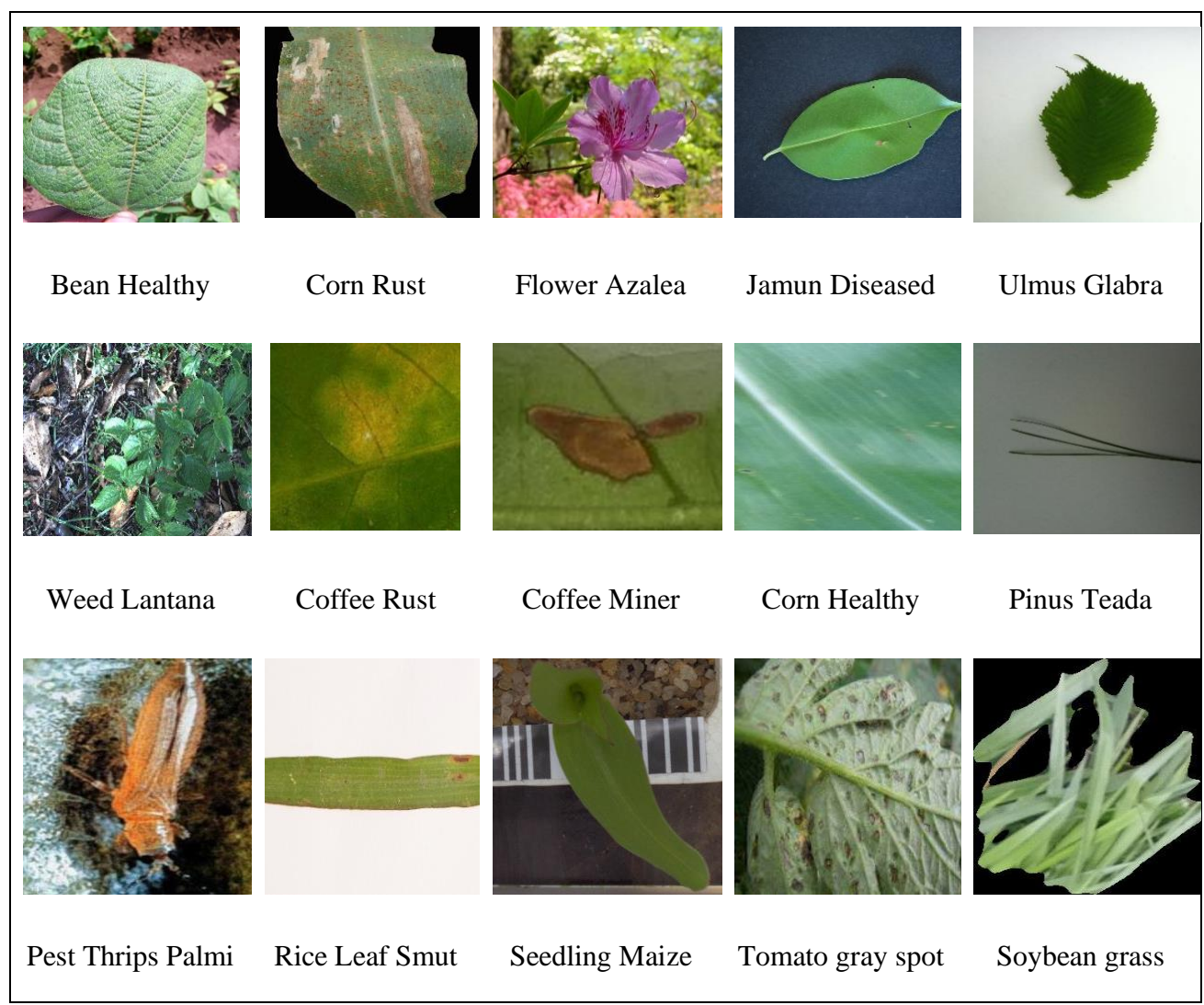

| Bean Healthy | Corn Rust | Flower Azalea | Jamun Diseased | Ulmus Glabra |
| Weed Lantana | Coffee Rust | Coffee Miner | Corn Healthy | Pinus Teada |
| Pest Thrips Palmi | Rice Leaf Smut | Seedling Maize | Tomato gray spot | Soybean grass |

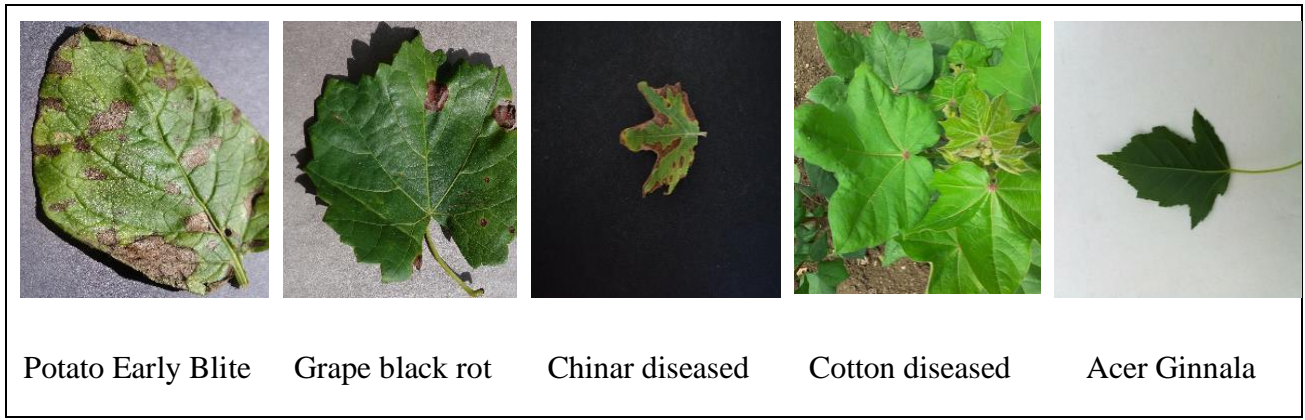

| Potato Early Blite | Grape black rot | Chinar diseased | Cotton diseased | Acer Ginnala |

**Figure 1 Sample images from AgriNet dataset**

The dataset classes were constructed by merging the same classes from multiple datasets in one class. This provides better classification performance through training the neural network to classify images regardless of the image location, quality, and device, which was a common challenge reported in the literature. For example, for the tomato plant diseases, images were combined from datasets [26,27,33] that included lab and field images from the United States, India, and Taiwan.

The collected dataset is highly imbalanced. As listed in Table 1, the average number of images per class is 378 images. Moreover, the number of classes with images less than 100 is 102 classes and the number of classes with images greater than 1000 is 44. In addition to class imbalance, a categorical imbalance between the three major tasks also exists. While training the models, the class weight mechanism was introduced to mitigate the class imbalance.

**Table 1 Summary of AgriNet dataset per category**

| Category | #Images | #Classes | Average | Median | Description | Reference |
|---|---|---|---|---|---|---|
| Species | 52150 | 309 | 169 | 144 | 12 mushroom, 103 flowers, 194 leaves | [15-121] |
| Pests & Weeds | 38305 | 33 | 1161 | 672 | 8 weeds,5 soybean weeds,8 pests,12 seedling | [22-25] |
| Diseases | 69687 | 81 | 1700 | 491 | 30 species | [24-33] |

## 2.2 Methods

### 2.2.1 Data Preprocessing

The dataset is a collection of images from multiple sources. All images were converted to JPEG format and resized to 224x224 pixels which is the size recommended for the deep learning architectures used. The dataset was then split into 70% train,10% validation, and 20% test. To increase the dataset size and ensure that the model is more robust in classifying images when visual effects are modified, image augmentation was applied to the training set. The augmented images were generated through varying brightness, rotation, width shift, height shift, vertical flip, zoom, and shear.

### 2.2.2 Convolution Neural Network

A ConvNet is a sequence of layers where in every layer of a ConvNet one volume of activations is transformed to another volume through a differentiable function [34]. Three main types of layers are stacked to build a ConvNet architecture: Convolutional Layer, Pooling Layer, and Fully-Connected Layer [34]. First, a Convolution layer computes the output of neurons that are connected to local regions in the input. Then, an activation function is applied, such as ReLU, which is max (0, x) thresholding at zero [34]. After that, a pooling layer performs down sampling operation along the spatial dimensions (width, height). Finally, the Fully-Connected layer is a classical neural network layer that computes the class scores [34].

### 2.2.3 Deep Learning Architectures

Deep learning architectures that were frequently used in agricultural research were selected to train the AgriNet dataset.

A-VGG16 and VGG19: VGG is named for the Visual Geometry Group at Oxford and was introduced by Karen Simonyan and Andrew Zisserman in 2014 [35]. The main contribution of this model was the usage of small-sized 3x3 convolutional filters. Pooling was done using Max-pooling over a 2 x 2-pixel window, with a stride of 2. VGG16 is the winning architecture of the ICLRLSVRC-2014 competition, having a top accuracy of 71.3% and a top-5 accuracy of 90.1% [35]. The model has a depth of 16 and 143 million parameters. The main difference between VGG16 and VGG19, which was ranked second in the competition, is the model depth which is 19 in VGG19 [35]. VGG19 achieved top accuracy of 71.3% and top 5 accuracies of 90% while having 138 million parameters [35].

B- Inception-v3: The inception model was introduced in 2012 by Szegedy et al. where the main contribution was "going deeper". The model proposed was 27 layers deep, including inception layers. The inception layer is a combination of a (1×1 Convolutional layer, 3×3 Convolutional layer, 5×5 Convolutional layer) with their output filter banks concatenated into a single output vector forming the input of the next stage [36]. Inception-v3 was introduced in 2016 as a convolutional neural network architecture from the Inception family with several improvements including usage of factorized 7 x 7 convolutions, label smoothing, and the use of an auxiliary classifier to propagate label information lower down the network [36]. Those improvements resulted in a top accuracy of 77.9% and a top5 accuracy of 93.7%. It is 159 layers deep and has 23 million parameters [36].

C-Xception: Xception model was proposed by Chollet et al. in 2017. It stands for "extreme inception" taking the principle of inception to an extreme. It is a convolutional neural network architecture that relies solely on depth-wise separable convolution layers [37,38]. The main difference between inception and Xception is that in inception, 1x1 convolutions were used to compress the original input, and from each of those input spaces different type of filters was used on each of the depth space. On the other hand, Xception reverses this step where filters are applied followed by compression. The second difference is the absence of non-linearities in Xception compared to the usage of ReLU in inception [37,38]. The Xception achieved a top accuracy 79% of and a top5 accuracy 94.5% of while having 22.9M parameters and a depth of 126.

D-InceptionResNetv2: InceptionResNetv2 was proposed by Szegedy et al. in 2016 and builds on the Inception family of architectures but incorporates residual connections by replacing the filter concatenation stage of the Inception architecture [39,40]. Residual connections allow shortcuts in the model leading to better performance while simplifying the Inception blocks [39,40]. The model achieved top accuracy of 80.3% and a top 5 accuracy of 95.4% while having 55.9M parameters and a depth of 572.

Thus, each of the used architecture has its benefits depending on the targeted applications. A detailed comparison of the models is presented in Table2. Note that Xception model has the smallest size of 88 MB while InceptionResNet-v2 achieved the highest top1-accuracy and top5-acuuracy of 0.803 and 0.953 respectively. In terms of depth and parameters, Xception model has the smallest number of parameters of 2291480 though VGG16 has the shortest depth of 23 layers.

**Table 2 Comparison of the ImageNet Architectures used in AgriNet**

| Model | Size | Top-1 Accuracy | Top-5 Accuracy | Parameters | Depth |
| --- | --- | --- | --- | --- | --- |
| Xception | **88 MB** | 0.790 | 0.945 | **22,910,480** | 126 |
| VGG19 | 549 MB | 0.713 | 0.900 | 143,667,240 | 26 |
| InceptionResNe-v2 | 215 MB | **0.803** | **0.953** | 55,873,736 | 572 |
| Inception-v3 | 92 MB | 0.779 | 0.937 | 23,851,784 | 159 |
| VGG16 | 528 MB | 0.713 | 0.901 | 138,357,544 | **23** |

### 2.2.4 Transfer Learning

Transfer learning is the state-of-the-art approach with scarce data applications. The common approach for vision-based application is to train a ConvNet on a very large dataset (for example, ImageNet, which contains 1.2 million images with 1000 categories), and then use the ConvNet either as an initialization or a fixed feature extractor for the task of interest [43]. Three Transfer Learning methods exist:

A-ConvNet as fixed feature extractor: This is done by removing the fully connected layer from the ConvNet pretrained on a generic dataset (ex. ImageNet), then treating the rest of the ConvNet as a fixed feature extractor for the new dataset.

B-Fine-tuning the ConvNet: The second approach adds to the first approach by fine-tuning the weights of the pretrained network by continuing the backpropagation [43]. Although retraining the whole model is possible, usually, some of the earlier layers are kept and we only fine-tune some higher-level portion of the network. This is because features of a ConvNet contain more generic features like edges in the first layers, but later layers of the ConvNet are more detail-oriented toward the pretrained model's classes [43].

C-Pretrained models: Final ConvNet checkpoints are frequently released to assist in fine-tuning tasks since modern ConvNets are time-consuming. For example, it takes 2-3 weeks to train a ConvNet across multiple GPUs on ImageNet [43].

First, transfer learning was applied on ImageNet pretrained models, where ImageNet was the generic dataset and AgriNet was the target dataset (Figure 2). After training the AgriNet models, the architectures with their weights were saved and proposed as pretrained models for any other agricultural classification task.

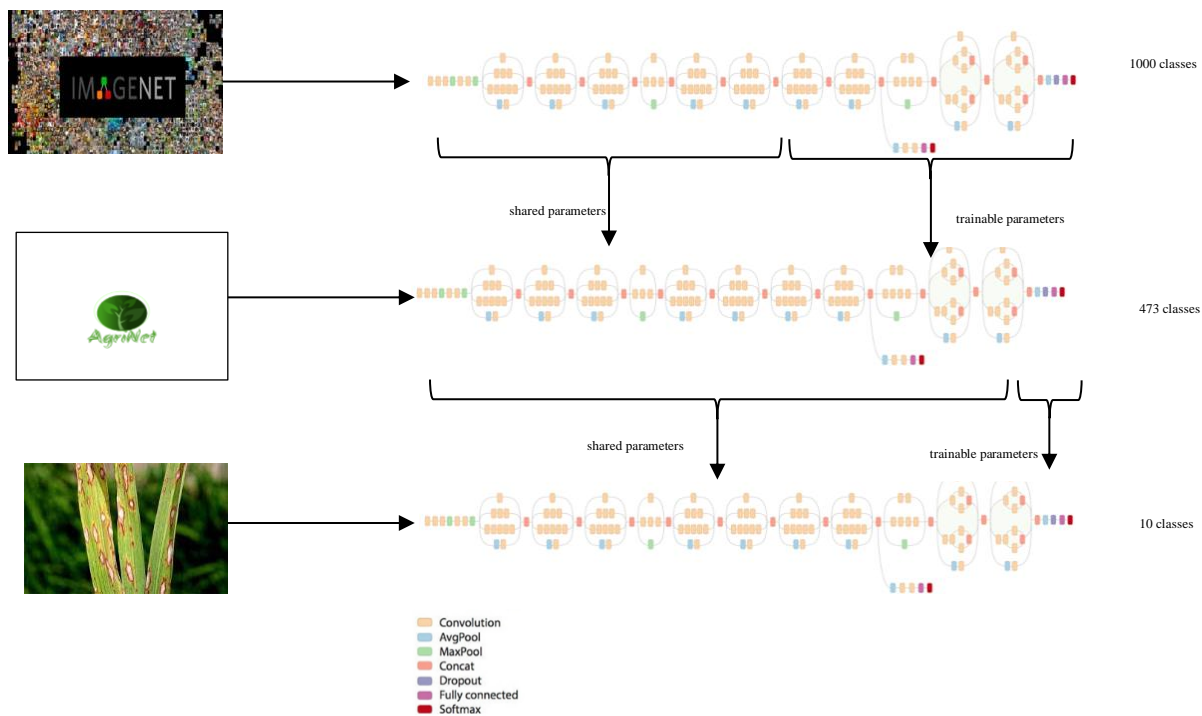

**Figure 2 Transfer Learning**

### 2.2.5 Improving the Models' Performance

To tackle the bias in the DL models, the severe class imbalance in the dataset, and to improve the models' convergence, three main methods were applied to the five AgriNet architectures:

A- Class weights: Class imbalance can affect the classification accuracy of small classes compared to large classes. To improve the performance of classification in small datasets, multiple solutions exist including oversampling, under-sampling, and class weight. In AgriNet, class weights were added to all the trained neural networks so that a balance is created between classes during the training process [44].

B-Decaying learning rate: Learning rate decay is a technique for training neural networks, by starting with a large learning rate and then decaying it multiple times [42]. It aims to improve optimization and generalization [45]. This improvement is an outcome of the fact that an initially large learning rate accelerates training or helps the network escape spurious local minima, and then decaying the learning rate helps the network converge to a local minimum and avoid oscillation [45].

C-Dropout: Dropout is a regularization method that approximates training a large number of neural networks with different architectures in parallel [46]. It was proposed by Srivastava et al. to resolve the overfitting problem in large DL models. Moreover, the term refers to dropping out units, which means temporarily removing units from the network, along with all its incoming and outgoing connections [46]. In the simplest case, each unit is retained with a fixed probability p independent of other units, where p can be chosen using a validation set or can simply be set at 0.5, which seems to be close to optimal for a wide range of networks and tasks. For the input units, however, the optimal probability of retention is usually closer to 1 than to 0.5[46].

### 2.2.6 Evaluation Metrics

Evaluation of the proposed models was based on two metrics.

A-Accuracy: accuracy represents the number of correctly classified data instances over the total number of data instances.

B-F1-score: The F1-score is the harmonic mean of precision and recall. Precision is the ratio of correctly predicted positive observations to the total predicted positive observations while recall is the ratio of correctly predicted positive observations to all observations in the actual class.

Accuracy is the most widely used metric to evaluate the performance of classification models. However, F1-score accompanies accuracy in classification tasks where the dataset is unbalanced.

## 3   Results and Discussion

### 3.1   Fine Tuning AgriNet Models

Transfer learning was applied to all AgriNet architectures. For each architecture, multiple experiments were done to propose the most accurate model by changing the number of frozen and trainable layers. The optimizer selected in VGG16 and VGG19 was SGD while in Inception-v3, Xception, and InceptionResNet-v2 Adam optimizer was used. All models were trained on a batch size of 32.

A-Inception-v3 experiments: The Inception-v3 model constitutes 311 layers. We tested freezing the first 133, 165, 197, 228, 249, and 280 layers. The layers are respectively named mixed4, mixed5, mixed6, mixed7, mixed8, and mixed 9. We found that freezing the first 165 layers (mixed5) achieved the highest performance.

B- Xception experiments: The Xception model has 132 layers. The model was trained while fixing the weights of the first 66, 76, 86, 96, 106, 116, and 126 layers. The layers are respectively named add_5, add_6, add_7, add_8, add_9, add_10, and add_11. We found that fixing the first 116 layers (add_10), achieved the optimum performance.

C-InceptionResNet-v2 experiments: The InceptionResNet-v2 model constitutes of 780 layers. We tested fixing the first 400,480,560,631, and 711 layers. The layers are respectively named

block17_8_mixed, block17_13_mixed, block17_18_mixed, block8_1_mixed, and block8_6_mixed. We found that the first 400 layers(block17_8_mixed) were frozen to achieve the most accurate classification.

D-VGG16: The VGG16 model has 19 layers. The model was trained while freezing the weights of the first 7,11,15, and 19 layers, which are respectively named layer to block2_pool, block3_pool, block4_pool, and block5_pool. We found that fixing the first 15 layers (block4_pool) achieved the highest performance.

E-VGG19: The VGG19 model constitutes 21 layers. We tested freezing the first 7,12,17, and 21 layers, which are respectively named block2_pool, block3_pool, block4_pool, and block5_pool. We found that the first 17 layers(block4_pool) were frozen to achieve the most accurate classification.

Thus, each of the architectures had its optimum freezing percentage. For Inception-v3 and InceptionResNet-v2, 53% and 51.3% freezing of weights achieved the highest accuracy respectively. On the other hand, for VGG16, VGG19, and Xception, the highest accuracies were achieved when freezing percentages of 78.9,77.3, and 87.9 respectively.

**Table 3 Fine-Tuning in AgriNet Models**

| AgriNet Architecture | Frozen Layers | Trainable Layers |
| --- | --- | --- |
| Inception-v3 | 165 | 146 |
| Xception | 116 | 16 |
| VGG19 | 17 | 5 |
| InceptionResNet-v2 | 400 | 380 |
| VGG16 | 15 | 4 |

## 3.2 Evaluation of AgriNet Models as Classification Models

### 3.2.1 Overall Networks Performance

After training the five architectures, the overall test and per class accuracies and F1-score were reported. Supplementary materials include the total number of images for each class and the per-class test accuracy for each of the five models.

**Table 4 AgriNet Models Evaluation**

|  | Train Accuracy | Val Accuracy | Test Accuracy | F1-score (macro average) |
| --- | --- | --- | --- | --- |
| Inception-v3 | 93.69 | 88.36 | 88 | 84 |
| Xception | 94.1 | 87.73 | 89 | 85 |
| VGG19 | **95.76** | **93.84** | **94** | **92** |
| InceptionResNet-v2 | 91.03 | 89.82 | 90 | 87 |
| VGG16 | 91.11 | 91.55 | 92 | 90 |

VGG19 surpassed all other models with a test accuracy of 94% and an F1 score of 92%. VGG16 was ranked second, followed by InceptionResNet-v2. However, the Inception-v3 model was the least performing with an average accuracy of 88% and an F1-score of 84% (Table1).

Another comparison was done for each of the models' sizes, the number of parameters, and Floating-Point Operations (FLOPs). InceptionResNet-v2 had the smallest FLOPs of 375,982,836 operations, followed by Xception with 623,900,414 operations and then VGG19 that reported 718,281,877 operations. Similarly, InceptionResnet-v2 had the smallest number of parameters which is 12,983,584 parameters, followed by VGG16 with 41,888,999 parameters and then by VGG19 with 94,092,935 parameters. For the smallest model size, VGG19 is ranked first with a model size of 159.9MB, followed by VGG16(180.1MB), and then by InceptionResNet-v2(980.3MB). Results are displayed in Table 5.

InceptionResNetv2 achieved the best compromise between accuracy, F1-score, FLOPs, number of parameters, and model size. Additionally, the model has a size of 980.3MB, the lowest FLOPs and number of t parameters, it achieved a 90% accuracy and an 87% F1 score.

**Table 5 Comparing ImageNet Architectures Used**

| | FLOPs | Total Number of Parameters-AgriNet | Size-AgriNet |
|---|---|---|---|
| VGG16 | 802,046,505 | 41,888,999 | **159.9 MB** |
| VGG19 | 718,281,877 | 94,092,935 | 180.1 MB |
| Xception | 623,900,414 | 124,056,527 | 1.28 GB |
| Inception-v3 | 951,318,693 | 74,666,183 | 816.2 MB |
| InceptionResNet-v2 | **375,982,836** | **12,983,584** | 980.3 MB |

### 3.2.2 Categorical Evaluation of the Models

The AgriNet Dataset consists of three main categories: species, weeds and pests, and diseases. Evaluation per category analysis was performed on each of the architectures proposed. VGG19 outperformed other architectures in species recognition and pests, weeds, and diseases detection tasks (Table 6).

**Table 6 Categorical Macro-average Test Accuracy**

| Category | Species | Pests and Weeds | Plant Diseases |
|---|---|---|---|
| VGG16 | 87.38 | 91.56 | 90.84 |
| VGG19 | **91.96** | **94.71** | **92.51** |
| Xception | 84.94 | 85.55 | 85.6 |
| Inception-v3 | 82.15 | 86.99 | 87.93 |
| InceptionResNet-v2 | 84.28 | 90.23 | 89.5 |

A-Species Classification Task

VGG19 achieved the highest accuracies in flowers, leaves, and mushrooms detection. Flowers classes are a combination of the VGG flowers dataset (103 classes) combined with the TensorFlow flowers dataset (5 classes merged with classes of the VGG flowers dataset). The VGG flowers dataset achieved a 70.4% accuracy [47]. As shown in Table 7, all AgriNet models outperformed the baseline model in the flower classification task. Similarly, leaves are majorly composed of the Leafsnap dataset which achieved an accuracy of 70.8 in [48]. In the mushrooms classification task, the highest accuracy of 75.77% was achieved by VGG19. The low accuracy of mushrooms classification compared to other classification tasks in AgriNet is mainly caused by the different image patterns of mushrooms compared to leaves and flowers constituting all other classes.

**Table 7 Macro Average Test Accuracies for Species Category on Test set**

| Category | #classes | Inception-v3 | Xception | VGG16 | VGG19 | InceptionResNet-v2 |
|---|---|---|---|---|---|---|
| Flowers | 103 | 80.63 | 86.92 | 86.88 | **93.47** | 80.63 |
| Leaves | 194 | 86.42 | 82.65 | 88.67 | **91.02** | 86.84 |
| Mushrooms | 12 | 69.83 | 72.44 | 70.71 | **75.77** | 73.7 |

B- Pests and Weeds Classification Task

The five pretrained models were able to achieve high accuracies in classifying pests and weeds as shown in Table 8. For weed images retrieved from the deep weeds dataset, the highest macro average accuracy of 89.9 was achieved by VGG19 while achieved the highest macro average accuracy of 74.93% using ResNet50 [50]. Same for weed seedlings, VGG19 achieved the top performance while reaching an accuracy of 98.62%. Moreover, for soybean weeds, all models achieved around 99% accuracy. Finally, in pests classification, Xception had the minimum accuracy of 94.87% and VGG19 had the highest accuracy of 98.62%.

**Table 8 Macro Average Test Accuracies for Pests and Weeds Category on Test set**

| Category | #classes | Inception-v3 | Xception | VGG16 | VGG19 | InceptionResNet-v2 |
|---|---|---|---|---|---|---|
| Weeds | 8 | 79.92 | 73.2 | 86.85 | 89.99 | 85.61 |
| Seedling | 12 | 76.72 | 81.88 | 88.66 | **92.72** | 86.35 |
| Pests | 8 | 94.95 | 94.87 | 95.74 | **98.62** | 94.9 |
| Soybean | 5 | 99.2 | 98.82 | 99.34 | **99.43** | **99.43** |

C-Plant Diseases Classification Task

Plant diseases classes were merged from different datasets. VGG19 was able to classify the largest number of plant diseases most accurately with a macro-average accuracy of 92.51%. VGG16 achieved top-class accuracy in a smaller number of classes than VGG19 and was ranked second with a macro-average accuracy of 90.84%. Sample macro average accuracies are presented in Table 9.

**Table 9 Macro Average Test Accuracies for Some Species Used in Plant Diseases Category on Test set**

| Category | #classes | Inception-v3 | Xception | VGG16 | VGG19 | InceptionResNet-v2 |
|---|---|---|---|---|---|---|
| Apple | 3 | 83.06 | 86.21 | 91.27 | **93.23** | 88.98 |
| Bean | 3 | 90.8 | 88.5 | 90.03 | **93.89** | 95.05 |
| Cassava | 5 | 67.77 | 64.1 | 71.3 | **72.75** | 69.29 |
| Cherry | 2 | 95.2 | 93.53 | 97.88 | **98.07** | 97.6 |
| Coffee | 5 | 94.14 | 92.82 | **97.52** | 97.26 | 95.67 |
| Corn | 4 | 93.07 | 92.23 | 93 | **95.51** | 90.68 |
| Cotton | 4 | 95.62 | 91.38 | **97.6** | 96.5 | 95.63 |
| Guava | 2 | 94.13 | 90.66 | 95.42 | **95.62** | 94.33 |
| Grape | 4 | 93.69 | 93.83 | **98.9** | 98.38 | 97.83 |
| Citrus | 2 | 90.3 | 98.09 | 93.87 | **99.59** | 89.57 |
| Mango | 2 | 86.8 | 91.5 | **98.11** | 96.64 | 93.34 |
| Potato | 2 | 87.1 | 88.28 | 94.73 | **95.20** | 85.17 |
| Rice | 3 | 83.33 | 75 | 83.33 | **87.5** | 75 |
| Strawberry | 2 | 95.05 | 97.97 | 97.75 | **99.54** | 99.54 |
| Tomato | 12 | 78.5 | 73.3 | 84.95 | **87.04** | 81.09 |

## 3.3 Evaluation of AgriNet Models as Pretrained Models

### 3.3.1 Evaluation on Rice Pests and Diseases Dataset

To evaluate the superiority of the proposed models, transfer learning was applied using ImageNet and AgriNet weights for the five ImageNet architectures on the rice pest and plant diseases dataset [51]. The dataset was split into 70% training, 10% validation, and 20% test sets. This dataset is a collection of 1426 field images of rice pests and diseases collected from paddy fields of Bangladesh Rice Research Institute (BRRI) for 7 months (Table 10).

AgriNet models achieved higher accuracies than ImageNet models. In VGG16, the model achieved a 90% accuracy using AgriNet weights, compared to 83% for ImageNet weights (Figure 3). On the VGG19 side, using AgriNet weights resulted in an 88% accuracy compared to 83% accuracy using ImageNet weights. Although VGG19 achieved the highest accuracy on the AgriNet dataset, VGG16 performed better on the rice pest and plant diseases dataset. This can be caused by specific image features that vary between a dataset and another. Thus, it is recommended to evaluate any agricultural dataset on multiple AgriNet architectures to achieve the best performance possible. It should be noted that the above accuracies on the dataset resulted after freezing the models and only training the dense layers. Further experiments can result in higher accuracies.

**Table 10 Rice Pest and Diseases Dataset Description**

| Class | Number of Images |
|---|---|
| False Smut | 93 |
| Brown Plant Hopper | 71 |
| Bacterial Leaf Blight | 138 |
| Neck Blast | 286 |
| Stemborer | 201 |
| Hispa | 73 |
| Sheath Blight and/or Sheath Rot | 219 |
| Brown Spot | 111 |

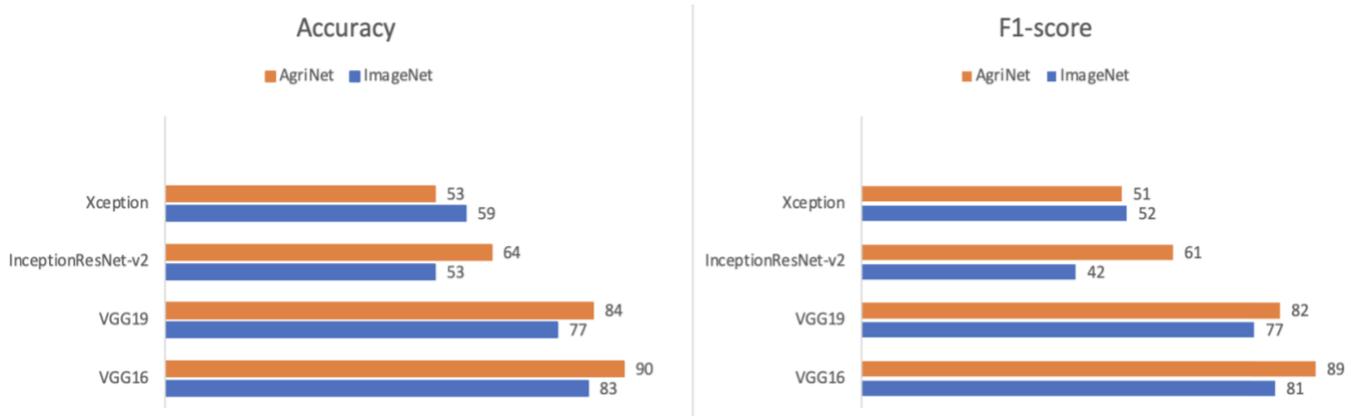

ΩΩ

**Figure 3 Models Comparison on Rice Pests and Diseases Dataset**

### 3.3.2 Evaluation on Plant Diseases of Kashmir Dataset

The plant diseases dataset of Kashmir contains 2136 images for eight plant species: Apple, Apricot, Cherry, Cranberry, Grapes, Peach, Pear, and Walnut with a total of 1201 healthy images and 935 diseased images (Table 11) [52]. The dataset was split into 70% training, 10% validation, and 20% test set. Similar to the case of the pest and plant diseases dataset, VGG16 achieved the highest accuracy and F1-score for both ImageNet and AgriNet models. All five AgriNet models achieved higher accuracies than ImageNet models. Moreover, the highest AgriNet accuracy reported was 83% compared to 70% in ImageNet on VGG16 as mentioned above (Figure 4).

**Table 11 Plant Diseases of Kashmir Dataset Description**

| Class | Number of Images |
| --- | --- |
| Apple Healthy | 93 |
| Apple Diseased | 100 |
| Apricot Healthy | 86 |
| Apricot Diseased | 100 |
| Cherry Healthy | 82 |
| Cherry Diseased | 95 |
| Cranberry Healthy | 100 |
| Cranberry Diseased | 94 |
| Grapes Healthy | 100 |
| Grapes Diseased | 9 |
| Peach Healthy | 100 |
| Peach Diseased | 18 |
| Pear Healthy | 100 |

| Pear Diseased | 58 |
|---|---|
| Walnut Healthy | 93 |
| Walnut Diseased | 100 |

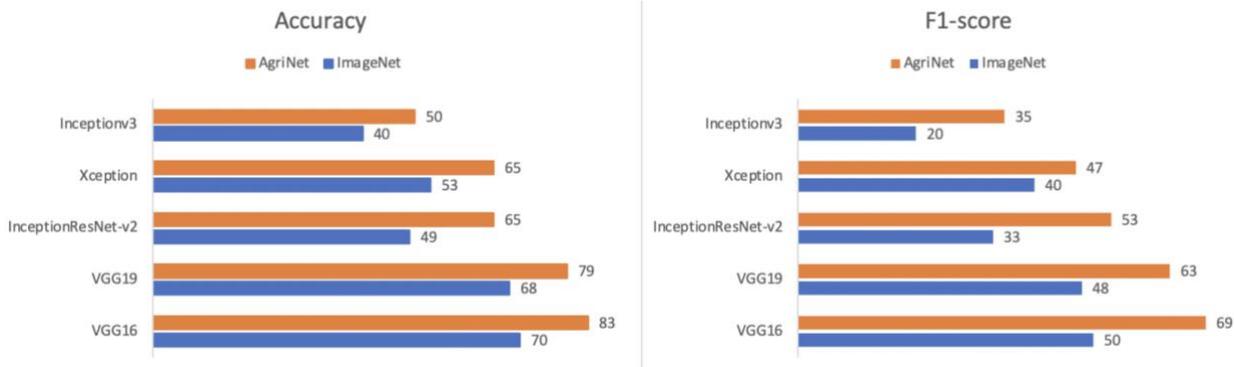

**Figure 4 Models Comparison on Plant Diseases of Kashmir Dataset**

## 4    Conclusion

In this paper, we present AgriNet dataset and AgriNet models, a collection 160k agriculture images dataset and a set of five agriculture-domain specific pretrained models respectively. VGG architectures achieved the highest accuracy with 94% accuracy for VGG19 and 92% accuracy for VGG16. InceptionResNet-v2 had the best compromise between the model's performance, and the computational cost through the number of trainable parameters, FLOPs, added to the model's size. In addition, the superiority of the proposed models was evaluated by comparing the AgriNet models with the original ImageNet models on two external pest and plant diseases datasets.VGG architectures resulted in best performance in both ImageNet and AgriNet models, where AgriNet surpassed the ImageNet models with accuracy increase of 18.6% and 8.4% using VGG16 for Kashmir dataset and rice dataset respectively.Further advancements to the AgriNet project include training the AgriNet dataset on more recent convolutional neural networks architectures, expanding pretraining to vision transformers, and increasing the dataset size through adding extra datasets or through applying advanced image augmentation techniques. However, adding additional datasets is restricted to the limited number of agricultural public datasets, which urges the research community in retrieving private datasets to public status whenever possible.

**Declaration of Competing Interest**

The authors declare that they have no known competing financial interests or personal relationships that could have appeared to influence the work reported in this paper.

**Availability of Data and Material**

The AgriNet dataset and the AgriNet pretrained models will be publicly available.

R E F E R E N C E S